\documentclass[journal]{IEEEtran}
\usepackage{subfigure}
\usepackage{url}
\usepackage{amsmath}
\usepackage{mathptmx}
\usepackage{mathtools}
\usepackage{multirow}
\usepackage{multicol}
\usepackage{changepage}
\usepackage{algorithm}
\usepackage{algorithmic}
\usepackage{balance}
\usepackage{graphicx}
\usepackage{tabularx}
\usepackage{tabulary}
\usepackage{footnote}
\usepackage{caption}
\usepackage{array}

\newcolumntype{C}[1]{>{\centering\let\newline\\\arraybackslash\hspace{0pt}}m{#1}}

\begin{document}


\title{Real-time 3D object proposal generation and classification under limited processing resources}

\author{Xuesong Li, Jose Guivant, Subhan Khan 

\thanks{X. Li (e-mail: xuesong.li@unsw.edu.au); J. Guivant (e-mail: j.guivant@unsw.edu.au); S. Khan (e-mail: subhan.khan@unsw.edu.au). All above authors are with the School of Mechanical Engineering, University of New South Wales, Sydney, NSW 2052, AU} }

\maketitle

\begin{abstract}
The task of detecting 3D objects is important to various robotic applications. The existing deep learning-based detection techniques have achieved impressive performance. However, these techniques are limited to run with a graphics processing unit (GPU) in a real-time environment. To achieve real-time 3D object detection with limited computational resources for robots, we propose an efficient detection method consisting of 3D proposal generation and classification. The proposal generation is mainly based on point segmentation, while the proposal classification is performed by a lightweight convolution neural network (CNN) model. To validate our method, KITTI datasets are utilized. The experimental results demonstrate the capability of proposed real-time 3D object detection method from the point cloud with a competitive performance of object recall and classification.

\end{abstract}

\begin{IEEEkeywords}
Point cloud segmentation, object detection, Optimization 
\end{IEEEkeywords}

\IEEEpeerreviewmaketitle

\section{\uppercase{Introduction}}\label{sec:introduction}

Robots are gradually undertaking more and more tasks in our daily lives, such as driving \cite{urmson2008autonomous} and providing service for hospitals \cite{ozkil2009service}. Introducing fully autonomous robots into our daily lives implies that robots share the spatial world with people. Due to this situation, it is imperative for robots to recognize all objects in the surrounding environments and understand the scenes they visualize \cite{geiger20133d}. Over the past ten years, approaches based on deep learning have achieved remarkable progress in vision-based object detection \cite{deng2009imagenet, visapp19}, and 3D object recognition from the point cloud \cite{dai2017scannet}, which makes the perception systems in robots more reliable and accurate than ever before. However, the deep learning model requires that a robot be equipped with energy-intensive processing capabilities, such as a GPU, for detecting 3D objects in real-time. Furthermore, a large number of small mobile robots are not able to dedicate those energy resources due to limitations in power supply and energy storage. Fortunately, certain relevant parts of the processing can be saved if proper hybrid approaches are exploited, which implies that power requirements can be reduced. In this work, we propose an efficient 3D object proposal generation and classification approach, which is capable of detecting objects in real-time with limited processing resources.

Three-dimensional object detection can provide accurate spatial location and geometrical shapes of targets, benefiting a robot's perception system. The traditional approach has been to segment out objects of interest (OOIs) and then apply some simple classifiers for further classification, such as a naive Bayes classifier \cite{nive_classifier} or support vector machine (SVM) classifier \cite{Hearst:1998:SVM:630302.630387}. With the prevalence of deep learning, many methods based on CNN have been proposed for 3D object detection \cite{3DSemanticSegmentationWithSubmanifoldSparseConvNet, s18103337, DBLP:journals/corr/LiZX16}. Despite their high detection accuracy, the deep learning-based methods require a large amount of annotated data for training purposes, and powerful processors for both training and real-time detection. However, many robots do not usually have powerful processing capabilities, especially small and medium-sized robots. When using these methods, a relevant part of the processing effort is dedicated to extracting point features, and generating 3D proposals. However, unlike 2D proposals in the image, 3D proposals are usually well scattered in 3D space. Taking advantage of this characteristic, we propose an efficient segmentation method for searching for 3D proposals, which reduces the computation effort significantly without sacrificing the quality of the estimated proposals. In addition, an efficient and lightweight deep learning model is developed for learning pointwise features for the classification task. Hence, our detection framework consists of two main parts: proposal generation and proposal classification.

The proposal generation firstly involves a pruning step for the removal of points estimated to be part of the ground. Because it is assumed that the OOIs are not part of the ground, the segmentation process can ignore all those points assumed to belong to that surface. The majority of the approaches used for estimating a road or ground surface assume that the surface is flat or possesses a similarly simple shape; consequently, these approaches attempting to approximate the surface with a plane. However, this assumption is not consistent with the real-world scenario, in which the ground/road contains several slopes within an area. Consequently, to achieve better estimation, the ground/road surface is assumed to be piece-wise multi-linear (PWML) \cite{PWML}, or at least piece-wise constant (PWC). In these cases, the area is divided into smaller but sufficiently large regions, in which dominant almost-horizontal, piece-wise approximating patches are assumed to be part of the ground/road surface in that region of the PWML or PWC partition. A dominant patch takes on the role of a local ground/road reference for classifying points as being part of saliences, depressions in the terrain, or the ground/road. 

The segmentation process is based on the continuity of the surfaces. Each inferred segment is considered as a potential proposal,  except for some improper segments, which are too large or too small. There are several parameters involved in the proposal generation step; these parameters are tuned once through a process based on particle swarm optimization (PSO) \cite{ps_optimization}. The efficient PointNet \cite{DBLP:journals/corr/QiSMG16, qi2017pointnetplusplus} used to perform the classification task consists mainly of multi-layer perceptron (MLP) on each point, which is equivalent to a 1 x1 x1 convolutional kernel in 3D, and it can perform the real-time classification on a standard CPU. Moreover, every proposal defines its own size; consequently, no bounding box regression task is required. Hence, the model is simpler and faster. In addition, it requires smaller training datasets than those needed for end-to-end CNN-based detection methods \cite{s18103337, avod_3d}. It is even possible to train the model with limited processing resources, such as those available in a common CPU.

The performance of the proposed method is evaluated using well-known KITTI datasets \cite{geiger2012we} via the consideration of three metrics: recall of proposals, the accuracy of the classification, and the processing time. Experimental results show that the method can attain high recall and accuracy in a real-time fashion and under the limited processing resources, typically found in small and medium-sized robots. This work makes two main contributions. The first contribution is the approach for inferring the ground, which is approximated via a PWC representation. Compared with other ground extraction method \cite{5979818, pfrsegmen, fast_segmentation} in the traffic context, this representation does not require the assumption that ground in the point cloud is one even plane, and is able to deal efficiently with uneven terrains. The second contribution is the proposed framework for detection, which can achieve real-time performance, e.g. by offering processing times of less than 0.1 seconds per massive point cloud, without utilizing a GPU. Consequently, this approach can be exploited by small robots equipped with simple CPUs with 1 or 2 cores, or by low-scale GPUs.

The rest of the paper is organized as follows. Section \ref{sec:Related_work} introduces the related work, and Section \ref{sec:methodology} illustrates how to build an efficient 3D object detection method. Additional procedures for model optimization are presented in Section \ref{sec:optimization}, while experiments involving the proposed method are described in Section \ref{sec:experiments}. Section \ref{sec:conclusion} concludes our work and summarizes its contributions.

\begin{figure*}[!tbp]
\begin{adjustwidth}{0.1 cm}{}
	\includegraphics[width = \linewidth]{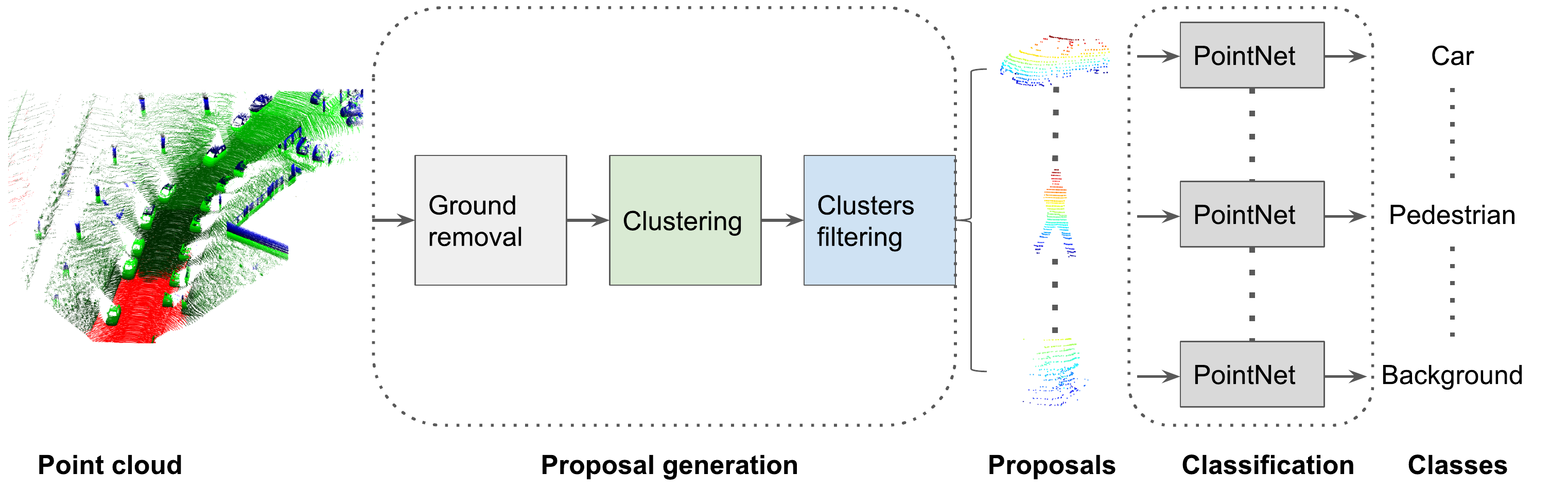} 
\end{adjustwidth}	
	\caption{The framework of our efficient 3D detection approach. Proposals are extracted from the raw point cloud with three procedures, ground removal, clustering, and cluster filtering. The PointNet is then employed to perform a classification for every cluster.}
\label{fig:framework}
\end{figure*}

\section{\uppercase{Related work}}\label{sec:Related_work}

The proposed 3D object detection framework can be split roughly into two parts: point cloud segmentation, and proposal classification. Related work concerning the point cloud segmentation will be firstly reviewed. Then a literature review of 3D object detection will be undertaken, to compare the proposed 3D object detection framework with those approaches.

\subsection{Point cloud segmentation}
Point cloud segmentation is used to infer objects of interest (OOIs) from a point cloud and is based on the assumption that 3D objects are sufficiently separated and there are no relevant intersections between adjacent objects. There are usually two main procedures involved: ground removal, and the clustering of the remaining set of points. Himmelsbach et al. \cite{Himmelsbach_2010} represented the terrain with a circle of infinite radius, centered around the robot itself, and split this circle into a number of circular sectors with fixed interval angles; Then the ground is searched by extracting horizontal lines at a low altitude, from the point sets of every segment. The remaining non-ground points are mapped into a 2D occupancy grid image, and clustering is done efficiently by finding connected components of occupied grid cells. Moosmann et al. \cite{graph_ground} constructed an undirected graph in which every point is treated as a node. The normal of the local surface is estimated for every node against its neighbouring points, and the similar local surfaces, which either form local convexities or are approximately vertical normal vectors, are merged together to grow the ground plane or obstacle area. 

Douillard et al. \cite{5979818} formulated the ground surface detection as non-ground outlier rejection and proposed a Gaussian Process Incremental Sample Consensus, based on the variance of the height of an outlier from the mean of a Gaussian distribution, to propagate the ground surface labels in an iterative manner to eventually include all ground points, while the remaining non-ground points are subsequently segmented into different clusters. Point segmentation approaches, after the removal of all ground points, often bring about issues of under-segmentation or over-segmentation. To alleviate segmentation errors, Held et al. \cite{pfrsegmen} built a probabilistic model to combine the spatial, temporal, and semantic cues for every segment. The probabilistic model iterates through every individual segment when conducting splitting and through all pairs of segments when merging. Zermas et al. \cite{fast_segmentation} started with an approximate coarse ground plane based on a deterministically extracted set of seed points with low height values, and iteratively refined the ground plane fitting by selecting seed points belonging to previously estimated ground surface. After removing the ground points, line segments are found efficiently for every scan and subsequently merged across neighbouring scans. One of the drawbacks in all these methods is that they include the assumption that one even plane should be able to fit the ground in the point cloud, but this assumption may fail in various real-life situations. In contrast, our ground removal approach adopts multiple small planes to fit the ground surface, resulting in a more robust approach.

\subsection{3D object detection}
The conventional pipeline in 3D object detection involves segmenting a point cloud, then classifying the clusters into objects. After point cloud segmentation, Teichman et al. \cite{5979636} adopted two classifiers, i.e., the segment and holistic classifiers, to classify feature vectors. The results were then combined with a discrete Bayes filter in a boosting framework. Wang et al. \cite{wang2015voting} proposed a feature-centric voting algorithm which convolutes the 3D point space in a sliding window with a voting scheme in a manner similar to feature transformation, enabling it to find new features describing object locations and orientations. An SVM classifier was subsequently applied to classify 3D candidates into different categories. As CNN-based approaches have become more and more dominant in computer vision tasks, a number of related 3D object detection approaches have been proposed. Chen et al. \cite{mutliviews} proposed multi-view 3D object detection and converted the point cloud into bird-view representations, consisting of multiple height maps, one density map, and one intensity map. The mature 2D CNN framework used for object detection in images was applied to these bird-view images to generate 3D object candidates. 

Ku et al. \cite{avod_3d} designed AVOD-FPN, which generates 3D proposals on the bird-view images of the point cloud with a 2D CNN, then crops the corresponding features from RGB and bird-view images for each proposal and predicts the 3D bounding boxes based on fused features. Zhou et la. \cite{DBLP:journals/corr/abs-1711-06396} adopted PointNet to extract features from the low resolution raw point cloud for each voxel, then employed a shallow 3D CNN network to convert 3D feature maps into 2D feature maps for 3D object detection. Other methods \cite{myownpaper, s18103337} use sparse CNN \cite{SubmanifoldSparseConvNet, DBLP:journals/corr/Graham15} to learn 3D feature maps from sparse 3D data, then compress them into 2D feature maps for object detection. Even though a large part of the computation for the sparse CNN is shifted to a CPU, these methods still require a GPU to achieve real-time performance. Although they can achieve desirable detection performance, these deep-learning-based approaches require a large training dataset and powerful processor, and cannot run in real-time without a GPU. To maintain both efficiency and accuracy, our framework employs point cloud segmentation to generate proposals and design an efficient classification model based on a CNN.

\section{Object detection methodology}\label{sec:methodology} 

The framework of the proposed detection approach can be seen in Fig. \ref{fig:framework}. It consists mainly of two parts: proposal generation and proposal classification. The proposal generation consists of ground removal, clustering, and cluster filtering. All of these steps are described in the following sections.

\begin{figure*}[h]
\begin{adjustwidth}{0.1 cm}{}
	\includegraphics[width = \linewidth]{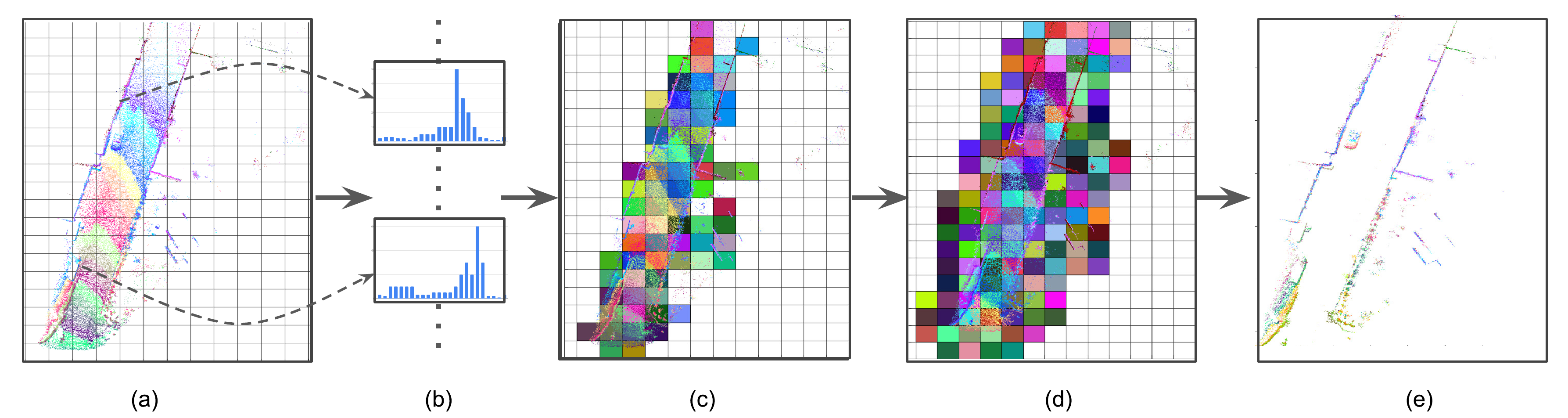} 
\end{adjustwidth}	
	\caption{The pipeline for the ground detection methodology: (a) a bird-view image of the entire original point cloud covering a full local area of size $80m$ by $70m$, every element of the partition is a fixed-size sub-region with size $4m$ by $3.5 m$; (b) the histogram of points' heights in every sub-region; (c) and (d) depict the detected ground planes before and after post-processing; (e) depicts the point cloud shown in (a) after the ground points are removed.}
\label{fig:ground_detection}
\end{figure*}

\subsection{Ground removal}\label{sec:ground_removal}

Pre-processing the data by removing points that belong to the ground has been shown empirically to improve segmentation performance significantly. The typical method employed in urban ground detection is to estimate an approximate 3D plane via model-fitting techniques, such as Random Sample Consensus (RANSAC). However, a simple plane is usually not sufficient for approximating the ground surface. A more powerful representation involves using a PWML approximation, such as in \cite{PWML}, or even more constrained cases, such as PWC with fixed partitions. The PWC assumes that in each region of the partition, the 'altitude' of the nominal ground is constant (i.e. the nominal surface of the terrain is flat and horizontal). In order to infer the altitude of the nominal surface in a subregion, a histogram of the altitudes of the points whose XY coordinates belong to this subregion can be generated. Additional basic rules, which consider the continuity of the properties of adjacent subregions, enable this process to avoid being misled by plateaus. The PWC requires very low processing effort, and performs better than the more sophisticated RANSAC.

Fig. \ref{fig:ground_detection} summarizes ground modelling based on a fixed-size PWC approximation. The coverage in XY ($W\times L$)  of the entire point cloud, is partitioned into multiple sub-regions, $W_{sub} \times L_{sub}$. For each subregion, a histogram of points' heights  (given simply by the points Z coordinates) is built. It is assumed that the ratio between the number of ground points and the number of non-ground points in every sub-region is larger than a certain user-defined threshold. According to this rule, the minimum bin value, whose frequency in the histogram is over this threshold, is assumed to be ground height. This operation can be implemented efficiently in parallel. The pseudo-code of the algorithm can be found in \ref{appendix:ground_removal}. Since the assumption may fail when a sub-region contains the flat plane belonging to an object, such as the roof of a car, we design the post-processing step to amend such false detections. The post-processing will compare the height of every sub-region with those of its adjacent regions, and its ground height value will be updated with the lowest height found in the neighbourhood. Once the ground planes are inferred, points can be classified as either being part of the ground ('close enough') or not. Based on this classification, only the points of interest (those above the ground) are kept for the posterior proposal generation of OOIs.

\subsection{Point cloud clustering}

The remaining points, i.e. the points 'over the ground', belong mostly to the OOIs, such as vehicles, cyclists, pedestrians, and background objects (buildings and plants). These objects are usually well separated when the ground points are removed. Therefore, we can perform clustering on these remaining points to find the potential 3D proposals.

Two methods are used to do point cloud clustering: distance-based and scan-based clustering. For the distance-based clustering method, the only criterion used to decide whether two points belong to the same cluster or not is their Euclidean distance. Hence, the distances between all point pairs are calculated and compared with a pre-defined distance threshold $T_d$. A random point $P_0$ is first selected and put into a separate list $C$. Then, the distances of other points from $P_0$ are calculated; Those points with distances from $P_0$ less than $T_d$, are removed and added to the same list $C$. Similar operations are performed repeatedly between points in $C$ and the rest of the points until no more points are added to list $C$, and $C$ is treated as a cluster. After this first cluster is complete, a new list will be created to search for the next cluster and so on until no more points are left. The algorithm details are shown in \ref{appendix:euclidean_distance}. Distance-based clustering, which does not require the ordering information among points, can be generalized to any point cloud dataset. However, processing a large number of point clouds takes a considerable amount of time, since every point is sequentially required to do a comparison against the rest of the points.

The 3D point cloud, generated by the LiDAR, has a multi-layer structure and every layer, also called a scan, produced from the same LiDAR ring includes contiguous points. Based on this characteristic, the scan-based clustering method is proposed to speed up processing. The points in every scan, or single layer, are first segmented into line segments, then these line segments are compared with their vertical neighbouring segments. Finally, the line segments which are close together are merged into one segment. 

It is assumed that points traverse in a raster in a counterclockwise fashion, starting from the top scan-line. $H_d$ is used to decide whether two points in the same scan belongs to the same segment or not, and $V_d$ is set to distinguish two points in neighbouring scans. We save all clusters, called $Global\_segs$, in the form of dictionary variables, in which the key and value are label number and corresponding points, respectively. All line segments begin with the first scan and are saved into a dictionary with the label as the key and the point group as the value, called $Above\_segs$. 

The first $Above\_segs$ is appended to $Global\_segs$. Then line segments are extracted from the second scan and saved in another dictionary, $Current\_segs$. Next, every line segment $seg\_i$ in $Current\_segs$ is compared with all segments in $Above\_segs$. If $Above\_segs$ has no segment whose distance from $seg\_i$ is smaller than $V_d$. Then, the key for $seg\_i$ in $Current\_segs$ is assigned a new label number, and this new key and its points are added to $Global\_segs$. If $Above\_segs$ includes one connection with $seg\_i$, no new key is generated, and the key for $seg\_i$ is the same as the one in $Above\_segs$; all points in $seg\_i$ are appended to the point group with the same key in the $Above\_segs$ and $Global\_segs$. If $Above\_segs$ includes more than two connections with $seg\_i$, the minimum key for the connected segments in $Above\_segs$ is selected as the unique key for these connected clusters.

Merging operations are implemented in the $Above\_segs$ and $Global\_segs$. After processing a current scan, the $Current\_segs$ becomes the $Above\_segs$ and the algorithm starts to process the next scan. The pseudo-code of scan-based clustering can be found in \ref{appendix:scan_based}. The author in \cite{fast_segmentation} designed a complicated label management system to handle label conflict and merge issues, while this paper employs the dictionary to save the points in all clusters directly, making our algorithm easier to implement. 

In contrast to the distance-based clustering method, the scan-based approach takes the advantage of the ordering of the points along the LIDAR scan-lines and covert clustering problem in 3D space into the line segmentation, and only the Euclidean distance between one point and its neighbouring points along the scan line is required, resulting in much lower algorithm complexity. Therefore, it can achieve a real-time running performance. But the scan-based approach is only applicable to point cloud which is generated by the Lidar or other sensors with fixed multi-layer scanning structure, and is not as generalized as the distance-based method.

\begin{figure}[!tbp]
\centering
	\includegraphics[width = 0.75 \linewidth]{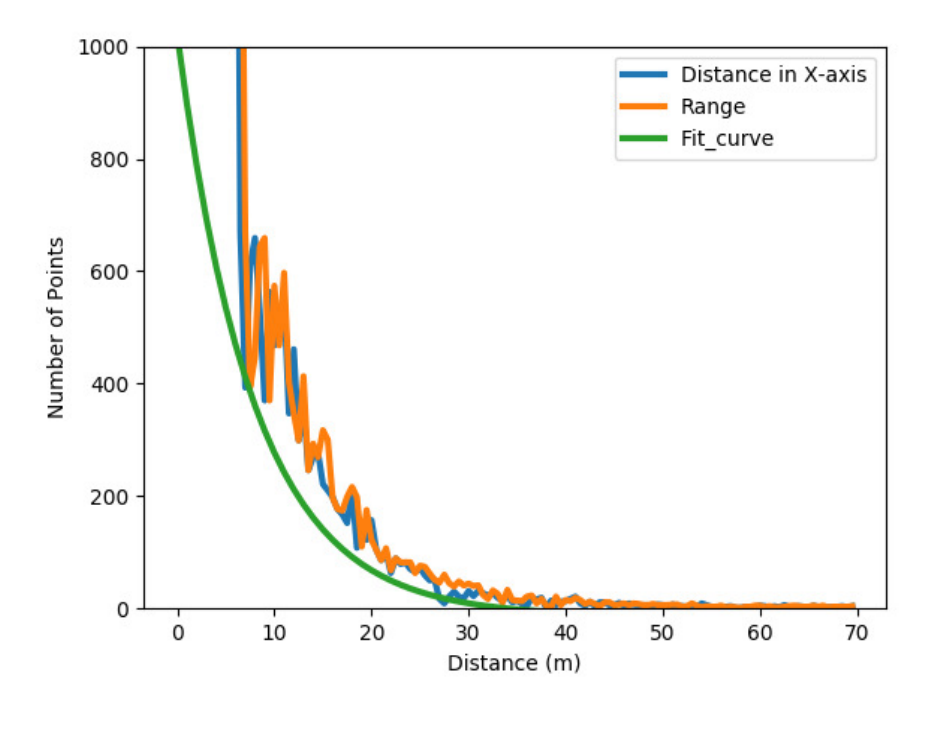} 
	\caption{The number of points within a proposal with respect to distance. The blue curve represents the number of points versus the distance along the X-axis to LiDAR, while the red curve represents the number of points against the range. The green curve is the fitted curve.}
\label{fig:occlusion}
\end{figure}

\subsection{Proposals filtering}

Three-dimensional proposals, generated by the clustering method, could also be for some background objects, such as super-large and tiny objects, which could be the buildings or leaves, respectively. To remove some improper background proposals and reduce the number of proposals, we set up two criteria to filter background proposals: 1) the size of the desired proposal should be within a certain range, and 2) the number of points in the non-occluded proposals should have a minimum. For the first criterion, we choose the maximum values for length and width to filter out very large objects, and the minimum value for height is defined to remove objects which are too flat, since objects of interest, such as pedestrians, cars, and cyclists, have certain height value.

As for the second criterion, the occlusion should be inferred first. The horizontal angle span of every 3D proposal is calculated against the location of the LiDAR and is compared with the angle spans of other proposals. A proposal is labelled as a non-occlusion proposal if it satisfies one of the following cases: 1) its angle scan does not overlap those of other proposals, and 2) its angle scan has overlapped with those of other proposals, but it is closer to the LiDAR than the other proposals. The details of this occlusion labelling algorithm are presented in \ref{appendix:occlusion_label}. To estimate the minimum number of points in non-occluded proposals, we investigate the relationship between the number of points and the distance from proposals to LiDAR, as shown in Fig. \ref{fig:occlusion}. The number of points and the distance to LiDAR are collected from all the ground truth bounding boxes in the training dataset provided by the KITTI benchmark \cite{geiger2012we}. The distance, which can be represented as either the range or the distance along X-axis to LiDAR, is sampled with an interval of $0.5$ $m$, and its corresponding number of points is the minimum number of points in all the ground truth bounding boxes, whose centre points are located in the distance interval. The exponential function is used to fit these sample points, see the green line in Fig. \ref{fig:occlusion}. Three-dimensional proposals with fewer than the minimum number of points are discarded.

\subsection{Proposals classification}\label{sec:proposal_classification}

\begin{figure}[!tbp]
\begin{adjustwidth}{-0.8 cm}{}
	\includegraphics[width = 1.1\linewidth]{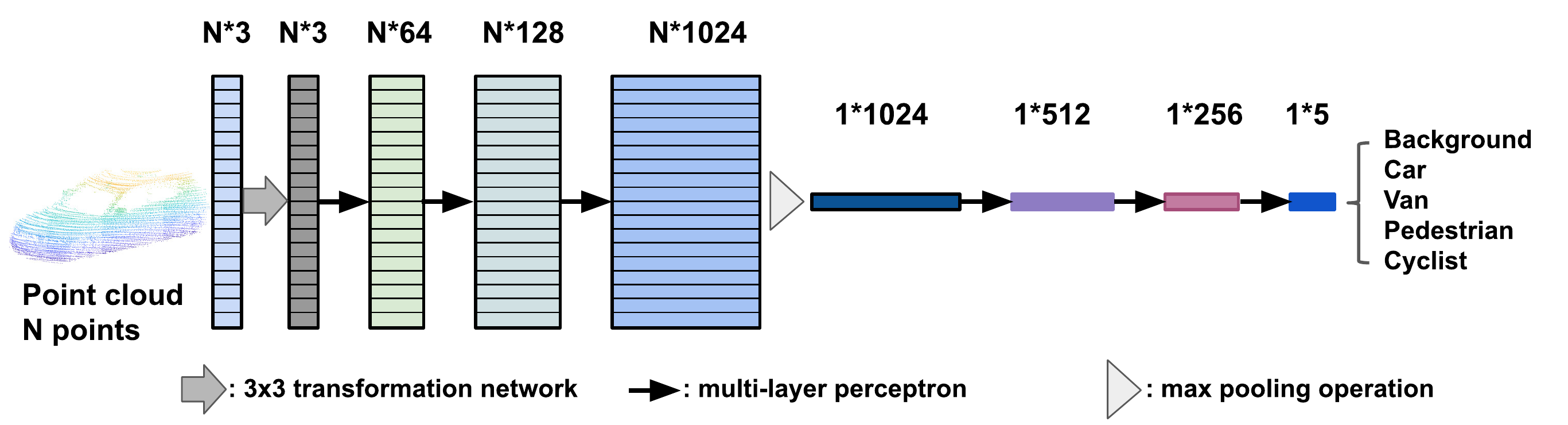} 
\end{adjustwidth}	
	\caption{Classification network. A MLP is applied on each point to increase the feature channel, and the parameters in the same layer is shared. Then max pooling is used to extract the global feature vector and output the final class scores}
\label{fig:pointnet}
\end{figure}

The PointNet-based, efficient, and lightweight classification model is designed to identify the class of 3D region proposals \cite{DBLP:journals/corr/QiSMG16}. There are numerous techniques available in the existing literature for processing raw point clouds in every proposal for classification, such as 3D CNN on a 3D voxelization grid \cite{myownpaper}, or a 2D CNN for bird-view images \cite{mutliviews}. In contrast to these methods, the PointNet can consume the raw points directly without altering the data representation, and it also employs an efficient and shared MLP to process the features on each point independently (i.e. pointwise $1\times 1\times 1$ CNN) and max-pooling operation, which can achieve real-time performances under constrained processing resources. Therefore, we choose the PointNet to build our proposal classification network, as shown in Fig. \ref{fig:pointnet}.

In terms of the classification model, the input consists of 3D coordinates $(x, y, z)$ along with $N$ points for one proposal. The raw point-cloud can be encapsulated int an $N\times 3$ matrix. The affine transformation on the point set (such as rotation) should not change the classification results for a geometric object. Therefore, a $3 \times 3$ transformation matrix generated by a spatial transformer network is first learned to transform the input points in order to make the model invariant to certain transformations. This type of transformation matrix is known as input 'TNet' \cite{DBLP:journals/corr/QiSMG16}. The PointNet in \cite{DBLP:journals/corr/QiSMG16} also include an $64 \times 64$ feature transformation matrix to align the features, but this matrix and corresponding feature 'TNet' is omitted in our classification model, as it increase a lot of computation with limited accuracy gain. To increase the channel features from $3$ to $1024$, three layers of MLP are applied with weights shared among all the $N$ point features. A max-pooling network is used to aggregate point features in order to generate $1 \times 1024$ global feature vector, and three layers of MLP are applied in order to decrease the channel features from $1024$ to $256$, then outputs the classification scores for $5$ classes: background, car, pedestrian, van, and cyclist.  In addition, batch normalization \cite{Ioffe:2015:BNA:3045118.3045167} is applied to all layers via ReLU \cite{Nair:2010:RLU:3104322.3104425} and dropout layers \cite{Srivastava:2014:DSW:2627435.2670313} are used for the fully-connected neural network.

\section{Parameters learning}\label{sec:optimization}

The proposed method consists of two sets of parameters: the parameters ($P$) in the proposal segmentation model, and the weights ($W$) in the classification model. To learn $P$, the PSO is applied, while gradient descent is employed to optimize $W$. This section will cover the optimization of these parameters.

\begin{figure}[ht]
\centering
	\includegraphics[width = 0.75\linewidth]{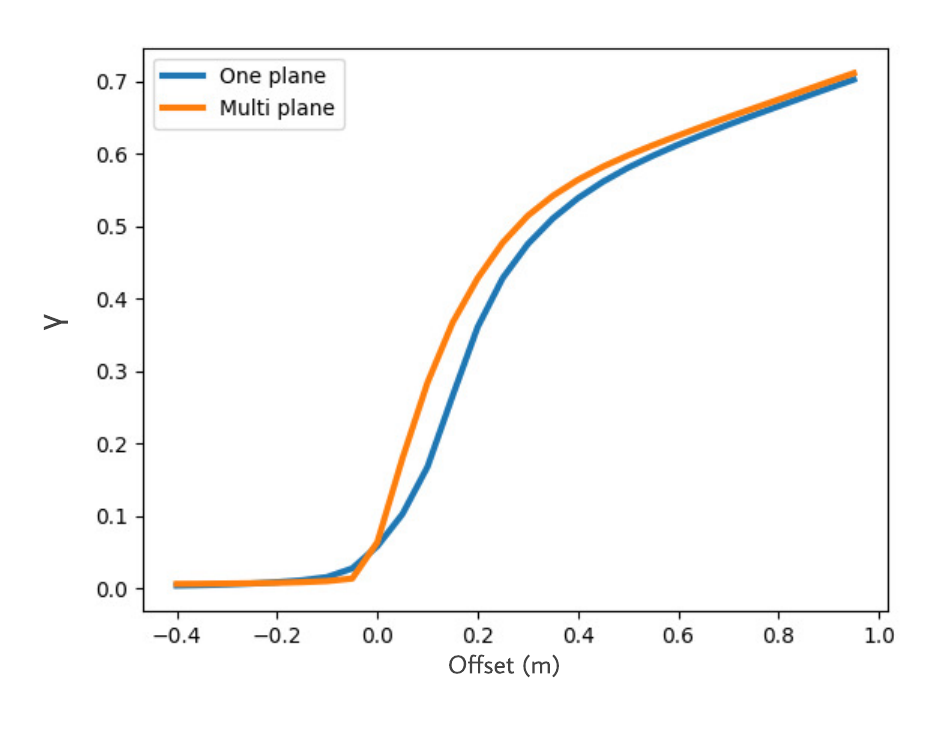} 
	\caption{The ratio between the number of removed points and all points versus the offset distance.}
\label{fig:ratio_r}
\end{figure}

\subsection{Proposal segmentation optimization}

The parameters in the proposal segmentation model include the distance threshold $H_d$ for the line segmentation done in one scan, merging threshold $V_d$ between adjacent scans, and filtering distance $D_o$ to the ground plane. The objective of the proposed segmentation method is to achieve the highest possible recall of objects via using PSO to optimize parameters. The PSO searches the space of an objective function by adjusting the trajectories of individual agents, called particles. Every particle, $x_{i}$, represents a solution, namely, $x_{i} = [H_d, V_d, D_o]$ in our case, and the objective function is the recall of objects, as shown in Equation (\ref{equ:recall}).
\begin{equation}
R = \frac{TP}{TP + FN},
\label{equ:recall}
\end{equation}
where $TP$ is the number of true positives, i.e. the correctly segmented objects, and $FN$ is the number of false negatives, i.e. the missed objects.

The movement of a swarming particle consists of two components: the stochastic and deterministic movements. Each particle is attracted towards the position of a current global best $g^{*}$ and its own best location $x^{*}_{i}$ in history, while, at the same time, it tends to move randomly. For each iteration, the particle $x^{t}_{i}$ is updated by Equations (\ref{equ:update}) and (\ref{equ:velocity}). The pseudo-code of PSO can be found in \ref{append:pso}.

\begin{equation}
x^{t+1}_{i} = x^{t}_{i} + v^{t+1}_{i}
\label{equ:update}
\end{equation}
\begin{equation}
v^{t+1}_{i} = \alpha\cdot v^{t}_{i} + \lambda\cdot r_{1}\cdot (g^{*} - x^{t}_{i}) + \theta\cdot r_{2}\cdot (x^{*}_{i} - x^{t}_{i}),
\label{equ:velocity}
\end{equation}

where $\alpha$, $\lambda$ and $\theta$ are acceleration constants for different terms, and $r_{1}$ and $r_{2}$ are the random number used to simulate diversity in the quality solutions.

\subsection{Classification model learning}
Considering that the number of labels in the different categories is balanced, we define a naive classification loss function, i.e. negative log-likelihood loss, with the following equation:

\begin{equation}
L_{cls}= - \frac{1}{N}\sum^{N}_{i=1}log(\hat{y}_{i}),
\label{equ:classification_loss}
\end{equation}
where $N$ is the number of training samples in one batch, and $\hat{y}_{i}$ is the predicted probability of $i$ belonging to the labelled class.

Since there exists a differentiable equation between the loss function (\ref{equ:classification_loss}) and parameters $W$, the gradient-based optimization method can be used to optimize the parameters $W$. The gradient descent method, i.e. AdamOptimizer \cite{adam}, is applied to do the optimization in our classification model with an initial learning rate of 0.0002 and an exponential decay factor of 0.8 for every 18570 steps.

\section{Experimental results}\label{sec:experiments}

\begin{figure*}[!tbp]
\begin{adjustwidth}{0.0 cm}{}
	\includegraphics[width = \linewidth]{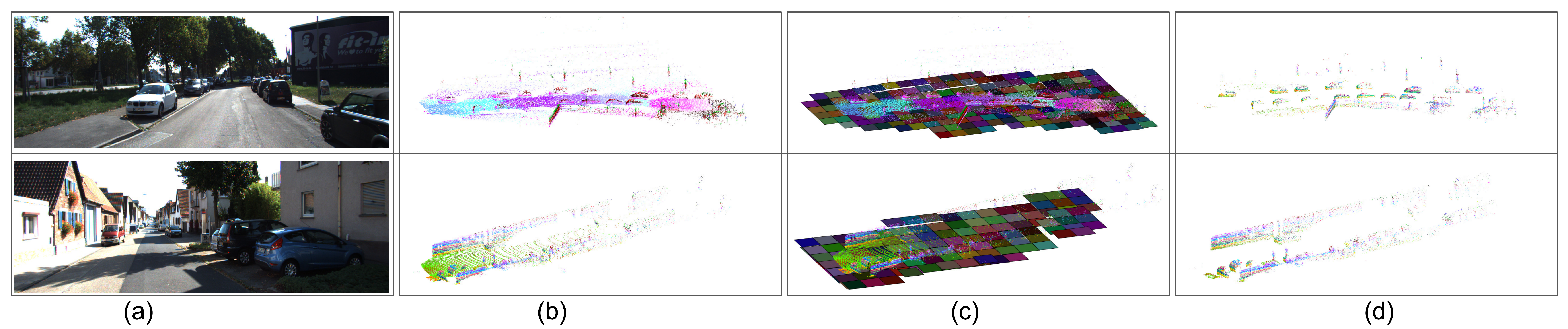} 
\end{adjustwidth}	
	\caption{ Visualization of the ground removal: (a) RGB image, (b) raw point cloud, (c) raw point cloud with ground planes, and (d) point cloud with ground points removed.}
\label{fig:ground_plane_removal_results}
\end{figure*}

\begin{figure*}[!h]
\centering
	\subfigure[The convergence graph of recall versus iteration.]{
	\includegraphics[width=0.33\textwidth]{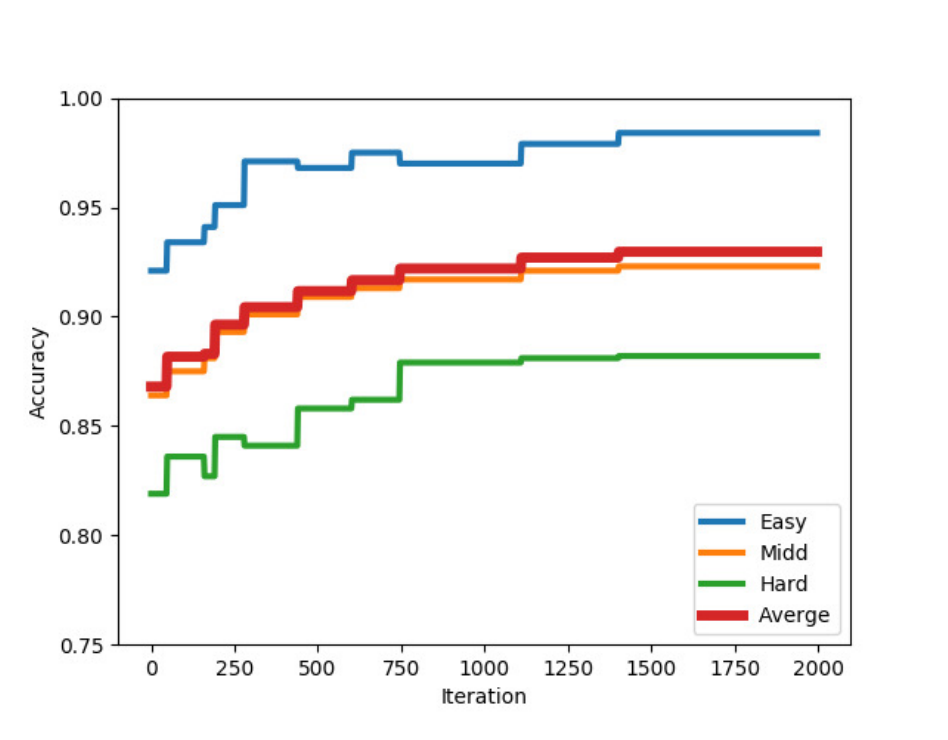}}
	\hspace{1.5cm}
	\subfigure[The convergence graph of the global particle versus iterations.]{
	\includegraphics[width=0.33\textwidth]{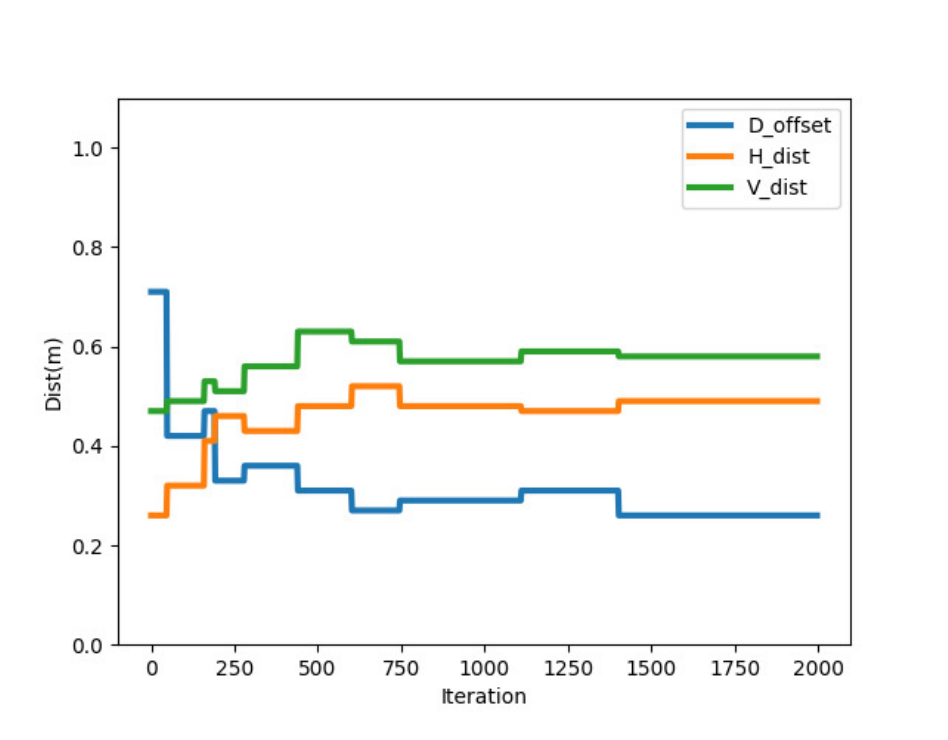}}
\caption{The convergence graphs for the PSO algorithm.}
\label{fig:pso}
\end{figure*}

\subsection{Implementation details}
\subsubsection{Dataset}
The KITTI benchmark dataset is adopted to validate our proposed method. It includes 7481 training pairs for the 3D LIDAR point cloud, RGB image, ground truth annotation, and calibration parameters. The optimization of the PSO and classification model training are conducted on this training dataset using different split ratios between the training and validation datasets.

\subsubsection{Detection method details}
The offset distance $D_o$ to the detected ground plane for removing ground points is set to $0.26$ $m$. For the task of clustering the point cloud, the $T_d$ in Euclidean-based clustering is assigned the value $0.5$ $m$, and the $H_d$ and $V_d$ in the scan-based method are also specified as $0.49$ and $0.58$ $m$, respectively. These parameters are optimized by the PSO. When calculating the objective fitness, i.e. recall value, the intersection over union (IoU) threshold for true positives is set to 0.25. For the ground removal, the user-defined ground ratio threshold for finding the ground height in every subregion is $0.05$, and the width of bin in histogram is $0.15$ $m$.


For the classification task, the number of points in one proposal is randomly sampled to 100 points, and all points are normalized by subtracting the central point from each other and then dividing this difference by the furthest distance from the center. All convolutional kernel sizes are $1\times 1\times 1$ in classification model, $3$ layers of CNNs are used to increase the channel number of features, and the corresponding feature maps sizes are $300 \times 64$, $300 \times 128$ and $300 \times 1024$. After the max-pooling operation, $3$ layers of fully connected networks compress the feature ($1\times 1024$ to $1\times 5$) through two middle features ($1 \times 512$ and $1 \times 256$).

\subsubsection{Hardware platform}
The processors in our hardware platform are 8 Intel(R) X-\\*eon(R) CPU E5-1620 v3 @ 3.50GHz, and every core has 2 threads. The memory capacity is 16 Gigabyte. The GPU card is a NVIDIA TITAN Xp with a 12 Gigabyte memory. Because this efficient detection method is proposed for robots with limited processing capabilities, we set an affinity mask to restrict our code to one core when generating the experimental data to make sure that our method can run well on small or medium robot platforms.

\subsection{Ground removal}

In order to compare our ground removal approach with R-\\*ANSAC qualitatively, we set up a criterion $\gamma$, i.e. the ratio between the number of removed points $N_g$ and the number of all points $N_a$, see Equation (\ref{equ:ratio_r}). The majority of points can be removed if the plane fits the real landscape of the ground plane. Therefore, given the same offset distance $D_o$ to the detected ground plane, the higher $\gamma$ is, the more effective the method is at removing the ground points. The qualitative experimental results and visualization of the ground removal performance can be found in Fig. \ref{fig:ratio_r} and \ref{fig:ground_plane_removal_results}. 

\begin{equation}
\gamma = \frac{N_g}{N_a}
\label{equ:ratio_r}
\end{equation}

The $\gamma$ is calculated by averaging all the ratio values of $1000$ point clouds randomly selected from the training dataset. From Fig. \ref{fig:ratio_r}, we can see that, when the offset distance is small, such as $0<D_o<0.3$ $m$, the $\gamma$ ratio for the proposed method is higher than that for RANSAC. This finding validates the effectiveness of the proposed method, as it can effectively remove more points on the ground surface and match the detected ground planes of the real landscape. In the case of a larger offset, it does not matter whether the detected plane fits the landscape closely or not. Thus, the ratios of the two methods converge towards each other.

\subsection{Proposal generation}

\begin{table}[ht]
\caption{The performance of the proposal method for different combinations.}
\label{table:clustering}
\centering
	\begin{tabular}{c|c|c|c}
	\hline
	Method modality & Recall(\%) & Number & Time (s) \\ \hline
	Clustering\_1+Filtering & 92.9 & 55 & 0.038 \\ \hline
	Clustering\_1& 94.0 & 146 & 0.032 \\ \hline
	Clustering\_2+Filtering & 93.3 & 78 & 2.351 \\ \hline
	Clustering\_2& 94.5 & 212 & 2.344 \\ \hline
	\multicolumn{4}{p{8.5cm}}{Clustering\_1 uses the scan-based clustering method, and clustering\_2 uses the distance-based approach (the time spent removing the ground is also included in clustering), the second column indicates the average recall value for each combination, and the third column lists the average number of proposals produced by each combination.}
	\end{tabular}
\end{table}

\begin{figure*}[h!]
    \centering
    \subfigure[Accuracy versus the number of input points.]{
    \includegraphics[width=.3\textwidth]{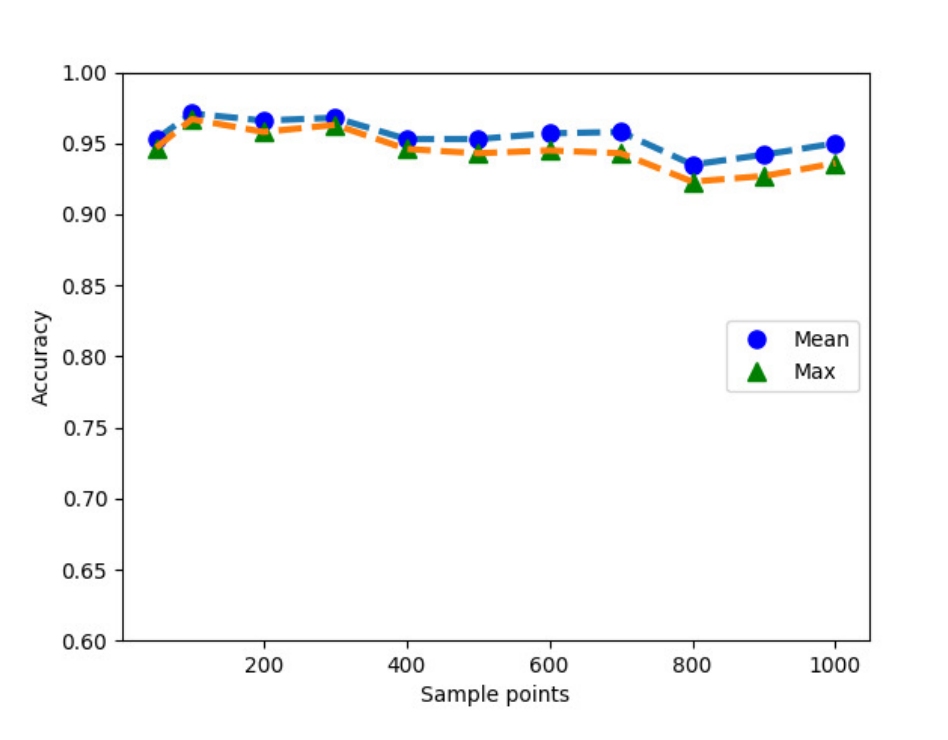}}
    \subfigure[mAP versus the number of input points.]{
    \centering
     \includegraphics[width=.3\textwidth]{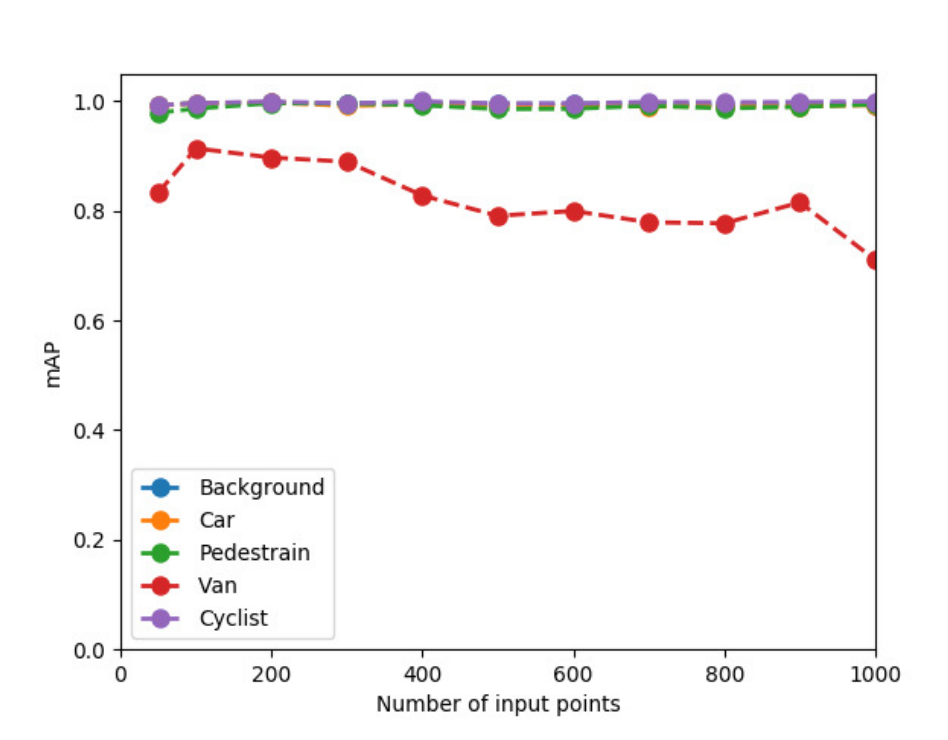}}
    \subfigure[Time versus the number of input points.]{
    \centering
    \includegraphics[width=.3\textwidth]{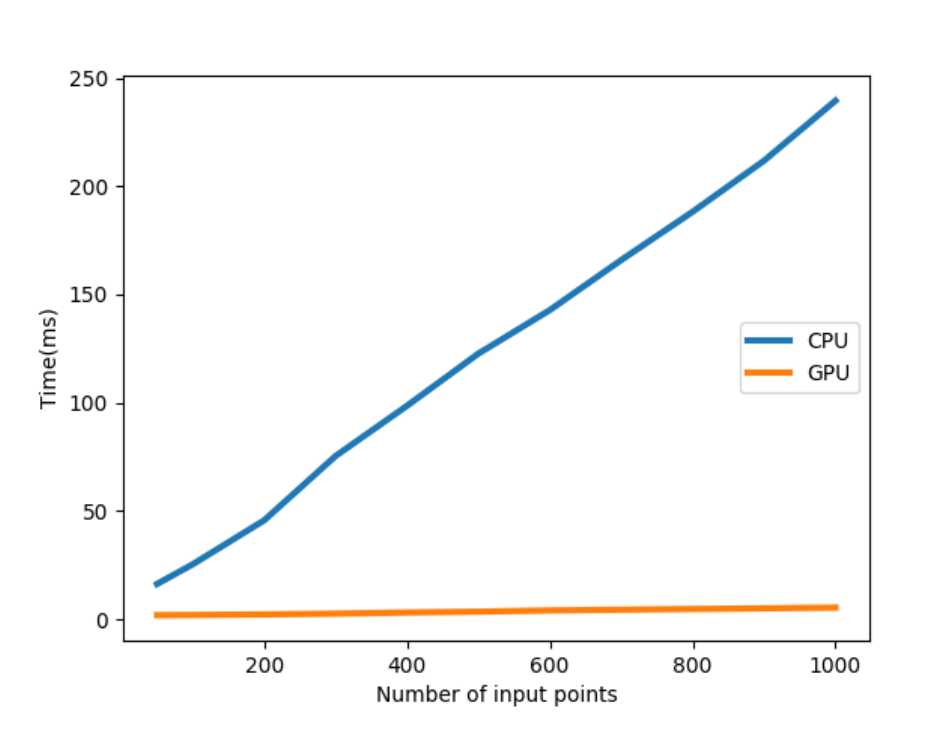}} \\
     \subfigure[Receiver operating characteristic curve.]{
    \centering
     \includegraphics[width=.3\textwidth]{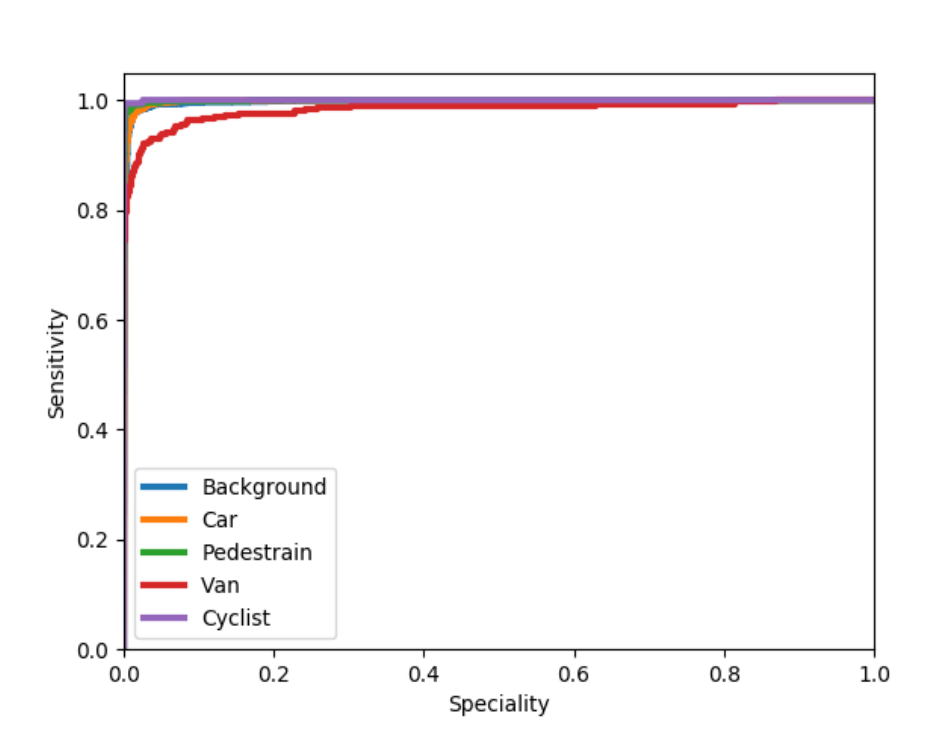}}
    \hspace{2pt}%
    \subfigure[Precision recall curve.]{
    \centering
    \includegraphics[width=.3\textwidth]{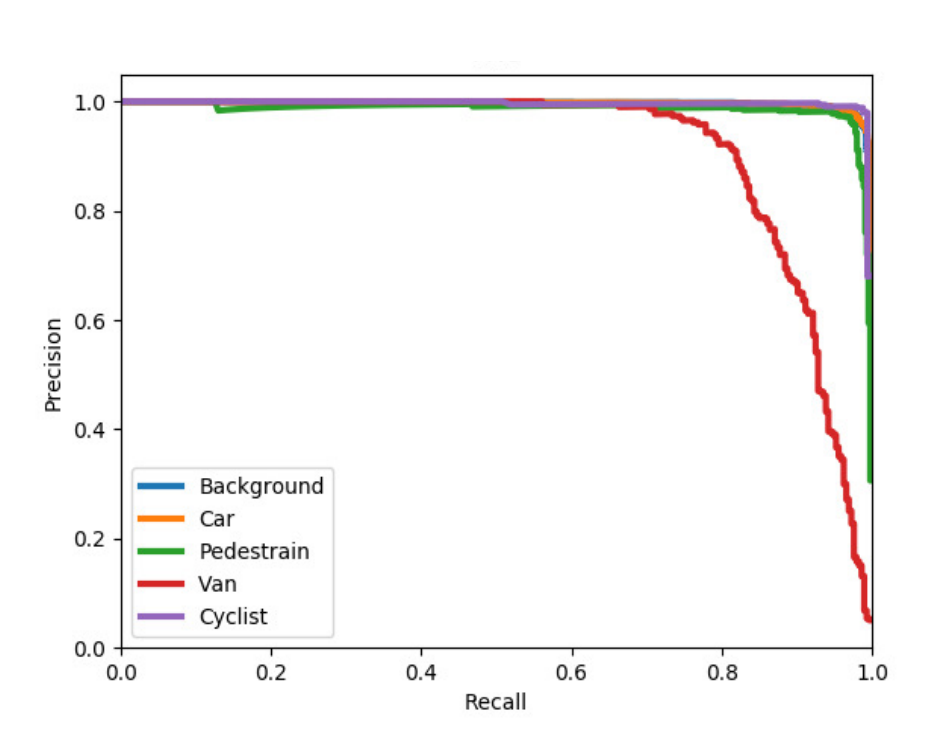}}
    \hspace{2pt}%
    \subfigure[The accuracy convergence of different split ratios of the training and testing data.]{
    \centering
    \includegraphics[width=.3\textwidth]{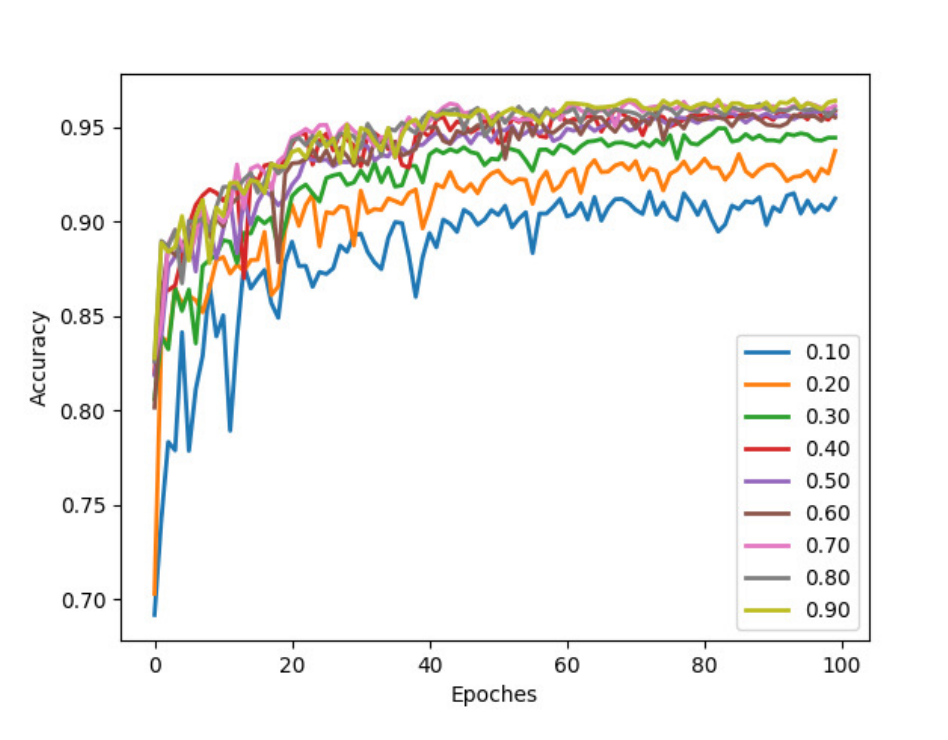}}  
    \caption{The analysis of the performance of the classification model. (a) Accuracy distribution ('Max' represents the best accuracy selected between 100 epochs, while the 'Mean' is calculated by averaging the accuracy of the last 50 epochs) versus the number of sample points. (b) The mAP of every class changes with the number of input points. (c) Running time performance on the CPU and GPU devices, given a batch size of 32. (d) The ROC plot for every category of detected objects. (e) The PRC plot for every detected class. (f) The split ratio of the training and testing data affects the speed of convergence and classification accuracy.}
\label{fig:classification}
\end{figure*}

In order to make experimental data, which is meaningful for robots with limited processing capabilities, we force our code into one core when running the experiments. Proposal generation consists of three steps: ground removal, clustering, and proposal filtering. For the scan-based clustering method, the processing times for these steps are $0.002$, $0.03$, and $0.006$ $s$, respectively. More experimental data is presented in Table \ref{table:clustering}, which shows that the scan-based method is much faster than the distance-based one. The distance-based method is slower because it calculates the distances from one point to all available points and searches a large number of adjacent points in each iteration. The performance of the proposal generation is very similar to that of the distance-based method, since both of them employ distance as the discriminative criterion regardless of the clustering mechanism used.

\begin{figure*}[!h]
	\includegraphics[width=1.0\textwidth]{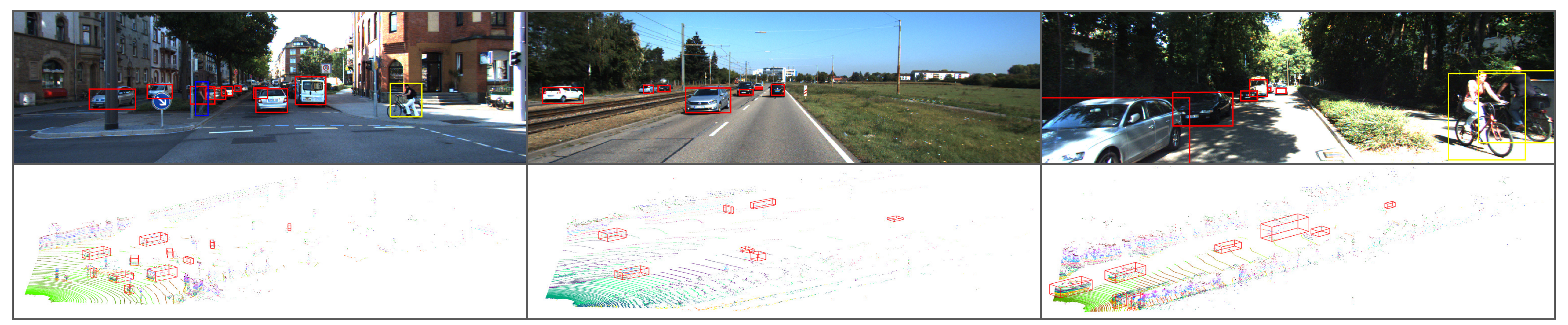} 
	\caption{Some samples of our detection approach. The bottom row shows the detection results in the point cloud, and the top row depicts the corresponding bounding boxes. The 2D bounding boxes in the images are generated by projecting 3D bounding boxes in Lidar coordinates into the image coordinates using the calibration parameters provided in the KITTI benchmark \cite{geiger2012we}. Here, the pedestrians are each denoted with a blue bounding box, the cyclists are each represented by a yellow bounding box, the cars are each in red bounding box.}
\label{fig:overall_perform}
\end{figure*}

The proposal generation method with clustering has a higher recall (\%) and three times the number of proposals compared to the combination of clustering and filtering, since the filtering procedure can filter out a number of false positives and reduce the number of proposals. However, it also removes some true positives and leads to a decrease in the recall value. In terms of the processing time, the clustering takes up to $0.032$ $s$, while the filtering only consumes $0.006$ $s$ and can decrease the number of proposals greatly. A large number of proposals will require a long processing time for the next classification task. Therefore, for a task involving limited computational resources, the filtering procedure is very important for achieving a real-time performance.

\begin{table*}[h]
\caption{The performance of the proposal method for different combinations against other methods.}
\label{table:overall}
	\begin{tabular}{C{4.0cm}|C{3.0cm}|C{2.0cm}|c|C{2.0cm}|C{2.0cm}}
	\hline
	\multirow{2}{*}{Method} & \multirow{2}{*}{Modality} & \multirow{2}{*}{Recall(\%)} & \multirow{2}{*}{Number of proposal} & \multicolumn{2}{|c}{Processing time (s)}\\
	\cline{5-6}&&&& GPU & CPU \\ \hline
	Clustering+Filtering+Classification & Lidar & 92.9 & 55 &  0.041 & $\textbf{0.082}^{\ast}$ \\ \hline
	Clustering+Classification& Lidar & 94.0 & 146 & \textbf{0.035} & $0.148^\ast$ \\ \hline
	AVOD-FPN \cite{avod_3d} & Lidar+Camera & 95.2 & 100 & 0.1 & 3.6 \\ \hline
	Voxelnet \cite{DBLP:journals/corr/abs-1711-06396}  & Lidar & -- & -- & 0.23 & 5.9 \\ \hline
	SECOND \cite{s18103337} & Lidar+Camera & -- & -- & 0.05 & 4.8 \\ \hline
	MV3D \cite{mutliviews} & Lidar+Camera & 94.4 & 50 & 0.36 & 11.6 \\ \hline
	\multicolumn{6}{p{18cm}}{The third column lists the average recall values at IoU threshold of 0.25 and the fourth column contains the average number of proposals produced by the method; the times spent on the GPU and CPU are the average times spent processing one point cloud; the time with \textbf{$^\ast$} is generated with only one core of CPU, otherwise, the 8 cores of CPU are used to report the time. The clustering method used is scan-based clustering. $'--'$ indicates unavailable data.}
	\end{tabular}
\end{table*}
 
To enhance the performance of the proposal generation me-\\*thod, $D_o$, $H_d$, and $V_d$, are optimized using the PSO algorithm. For every generation in the PSO, the recall value is calculated by running the proposal generation algorithm through $500$ training samples randomly selected from the entire training dataset. A total of 1000 generation and 50 particles are set for optimization. The convergences of the recall value and all parameters are illustrated in Fig. \ref{fig:pso}. All parameters are randomly initialized between $0$ and $1.2$ $m$. From Fig. \ref{fig:pso} (b), we can see that the parameters fluctuate at the beginning, then slowly converge to their optimal values. Since these parameters are constrained by each other, the combination of parameter values will decide the final fitness value. All parameter convergences fluctuate instead of moving in one direction. The convergence graph shows that the best values for $D_o$, $H_d$, and $V_d$ are $0.26$, $0.49$, and $0.58$ $m$, respectively.

\subsection{Classification model}

To reduce the overfitting and generalize the model, two basic techniques for data augmentation are introduced:
\paragraph{Rotation} Every point cloud in the entire block is rotated along the Z-axis and about the origin [0,0,0] by $\theta$, where $\theta$ is drawn from the uniform distribution [-$\pi$/4, $\pi$/4]. 

\paragraph{Scaling} Similar to the global scaling operation used in image-based data augmentation, the whole point cloud block is zoomed in or out. The $XYZ$ coordinates of the points in the whole block are multiplied by a random variable drawn from the uniform distribution [$0.95$, $1.05$].

We have investigated how the number of sample points $N$ in a proposal affects the classification accuracy and demonstrate their relationship in Fig. \ref{fig:classification} (a). It can be observed that the best accuracies of 0.971 (Maximum) and 0.967 (Mean) are achieved when $N=$100 for one proposal. The classification performance could deteriorate if more points (such as 1000 or more) are added to a proposal. The training samples contains 18000 background objects, i.e. 14357 cars, 1297 vans, 734 cyclists, and 2207 pedestrians, and Fig. \ref{fig:classification} (b), (d), and, (e) show the mean Average Precision (mAP) curve, the Receiver Operating Characteristic (ROC) curve, and the Precision-Recall Curve (PRC), respectively, for these samples. These curves demonstrate the detailed classification performance of our model for every class. It can be seen that the classification performance for vans is worse than those for other classes. Furthermore, the scores for the background, car, pedestrian, and cyclist are similar. The shape of the van is similar to that of the car, while the number of vans in the training sample is much smaller than the number of the car. Therefore, the classification model for identifying the van cannot be well-trained, leading to the poor classification performance for the van class. As for the cyclist class, even though the number of cyclists in the training sample is small, the shape of this class is distinct from those of other classes, which makes it easy for the classification model to capture its features. Thus, the model achieves a good classification performance for the cyclist class.

The running time for processing $32$ proposals versus different numbers of input points can be found in Fig \ref{fig:classification} (c). The implementation time for the CPU increases almost linearly with the number of sample points, while the GPU time is almost constant around $0.003$ $s$ due to the parallel computations. We find that it takes $0.0254$ $s$ to process $32$ proposals with $100$ points, making it feasible to run our classification model on the CPU-only hardware platform in real-time.

To capture how many data points are needed to train our model, we split the dataset into training and testing sets using different ratios, and the results can be found in Fig. \ref{fig:classification} (f). For this experiment, the number of points per proposal is 100, and we find that the classification performances are similar when the split ratio is larger than 0.3. Other end-to-end training detection methods, such as SECOND \cite{s18103337} and 3DBN \cite{myownpaper}, require a minimum 0.5 split ratio to guarantee consistent performance. In comparison, our framework for detection required less training data and had better capability in generalization because most of the background is removed by the segmentation algorithm, and most of the noise is peeled off, making it easier for this classification model to learn different classes.

\subsection{Overall performance}

To evaluate the performance, we assemble two sub-modules, i.e. proposal generation and classification, together into an object detection method. In Fig. \ref{fig:overall_perform}, the overall detection performance is presented\footnote[1]{More visuliazaiton results can be found in vedios: \url{https://youtu.be/poSbDQ1LCR0}, \url{https://youtu.be/7C4kOnLrtig}, \url{https://youtu.be/QhXHMC4wsMc}, \url{https://youtu.be/LrXg3J_4FKY}}. It can be observed that we can find bounding boxes which fit nearby objects well, but the bounding boxes are much smaller than the real sizes of the objects that are far away from the Lidar. For instance, in the top middle image, the 2D bounding boxes for cars, which are far away, only cover the bottom halves of these cars because the density of points on the surface of the object is inversely proportional to the distance between the object and Lidar. The proposal of a nearby object includes a number of points and these points can represent its real shape, while, for a far object, there are only a small number of points and only a part of the object can be captured.

The comparison of the computational efficiency of our met-\\*hod with other 3D detection methods is provided in Table \ref{table:overall}, which shows that the proposed method is much faster than other existing methods and can achieve a real-time performance on either a GPU ($0.035$ $s$) or CPU ($0.082$ $s$) hardware platform. As for the performance of the object proposal, AVOD-FPN \cite{avod_3d} and MV3D \cite{mutliviews} can obtain better recall value than ours, but their efficiency is very low, especially on the CPU platform, since they employ the CNN backbone networks to extract features images for generating proposals.

\section{Conclusion}\label{sec:conclusion}
In this paper, we have designed an efficient 3D object detection approach for robots with limited computational resources consisting of proposal generation and a lightweight classification model. The existing deep learning-based 3D object detection methods can achieve the desired detection performance but require the robots to equipped with a powerful processing unit, which is impossible in many small robots. In comparison to existing deep-learning techniques, the proposed method can achieve competitive recall values and classification accuracies. Most conspicuously, the proposed method has the capability of detection in real-time with either a GPU or CPU. In future work, we intend to fuse our work with a semantic segmentation task and combine the information from different resources for some complicated tasks, such as instance segmentation.

\section{Acknowledgment}
The presented research was supported by the China Scholarship Council (Grant No. 201606950019).



\bibliography{mybibfile}

\appendix

\section{Ground removal algorithm}\label{appendix:ground_removal}
\begin{algorithm}[H]
\begin{adjustwidth}{-0.2cm}{}
\caption{Output ground planes $Ground\_post$}
\label{alg:ground_removal}
\begin{algorithmic}
\STATE Legend:
\STATE
\begin{tabular}{p{0.8cm}p{7cm}}
$W$: &the width of whole region; \\
$L$: &the length of whole region; \\
$w_{sub}$: &the width of sub region; \\
$l_{sub}$: &the height of sub region; \\
$points$: &all points in the format [x, y, z] with the size $n \times 3$; \\
$ground\_num$:\\ &the minimum number of points in ground bin; \\
$bin\_width$:\\ &the width of bin in the histogram; \\
$subregion\_func()$: \\ &the function of getting all points in given sub\_region; \\
$histogram\_build()$:\\ &the function to build histogram; \\
$neighbourhood\_height\_func$: \\ & the function to compare the height in neighbourhood and return the lowest height value;\\
\hline
\end{tabular}
\STATE $Ground = array(w_{sub},l_{sub})$;
\STATE $Ground[:,:]=1000$;
\FOR{$i=1:ceil(\frac{W}{w_{sub}})$}
\FOR{$j=1:ceil(\frac{L}{l_{sub}})$}
    \STATE $points_{sub} = subregion\_func(points,i,w_{sub},j,l_{sub})$;
    \STATE $hist = histogram\_build(points_{sub}[:,2], bin\_width)$;
    \STATE $frequency_z = hist.frequency$;
    \STATE $bin_z = hist.bin$;
    \FOR{$k=1:len(frequency\_z)$}
        \IF{$frequency\_z[k] >= ground\_num$}
            \STATE $Ground[i,j] =height$;
            \STATE $break$;
        \ENDIF
    \ENDFOR
\ENDFOR
\ENDFOR
\STATE $Ground\_post = array(w_{sub},l_{sub})$;
\STATE $Ground\_post[:,:]=1000$;
\FOR{$i=1:ceil(\frac{W}{w_{sub}})$}
\FOR{$j=1:ceil(\frac{L}{l_{sub}})$}
    \STATE $low\_height = neighbourhood\_height\_func(i,j,Ground)$
    \STATE $Ground\_post[i,j] = low\_height$;
\ENDFOR
\ENDFOR

\end{algorithmic}
\end{adjustwidth}
\end{algorithm}

\section{3D point cloud clustering based on Euclidean distance}\label{appendix:euclidean_distance}
\begin{algorithm}[H]
 \begin{adjustwidth}{-0.2cm}{}
\caption{Output clustering results $Cluster$}
\label{alg:euclidean_distance}
\begin{algorithmic}
\STATE Legend:
\STATE
\begin{tabular}{p{1.3cm}p{7cm}}
$T_d$: &the minimum distance threshold; \\
$points$: &all the point cloud, $n\times 3$; \\
$dist\_func$: &distance calculation, $n\times n$; \\
\hline
\end{tabular}
\STATE $Clusters \leftarrow \{\}$: create empty cluster
\STATE $\textit{\textbf{D}}_{m} \leftarrow dist\_func(points)$
\WHILE{$\textit{\textbf{D}}_{m}$ is not empty}
    \STATE $stop \leftarrow False$
    \STATE $C \leftarrow {[]}$
    \STATE $C$ append $1$
    \STATE $index\_new \leftarrow 1$
    \WHILE{not stop}
        \STATE $index\_t \leftarrow fun\_index\_dist(index\_new, T_d,\textit{\textbf{D}}_{m})$
        \STATE $index\_t \leftarrow index\_t - C$
        \STATE $index\_new \leftarrow index\_t$
        \IF{ $index\_new$ is empty}
            \STATE $stop \leftarrow True$
        \ELSE
            \STATE $C$ append $index\_t$
        \ENDIF
        \ENDWHILE
    \STATE $Cluster \leftarrow get\_xyz(C, points)$
    \STATE delete the index\_t in points
    \STATE delete the index\_t in ${\textbf{D}}_{m}$
\ENDWHILE
\end{algorithmic}
\end{adjustwidth}
\end{algorithm}

\section{3D point cloud clustering based on scan order}\label{appendix:scan_based}
\begin{algorithm}[H]
 \begin{adjustwidth}{-0.2cm}{}
\caption{Output clustering results $Cluster$}
\label{alg:scan_based}
\begin{algorithmic}
\STATE Legend:
\STATE
\begin{tabular}{p{0.8cm}p{7cm}}
$H_d$: &minimum distance threshold insider one scan; \\
$V_d$: &minimum distance threshold between two scans; \\
$points$: &all the point cloud, $n\times 3$; \\
$mini\_points$:\\ &minimum number of connected points between two segments; \\
$find\_segments()$: \\ &find all segments in one scan with the distance threshold $H_d$; \\
$dist\_func()$:\\ &distance calculation between two segments; \\
\hline
\end{tabular}
\STATE $Clusters \leftarrow []$: create empty cluster 
\STATE $scans \leftarrow find\_all\_scans(points)$: rearrange the point cloud in the form of Lidar scan. 
\STATE $Global\_segs \leftarrow \{\}$: dictionary for segments in global setting
\STATE $Above\_segs \leftarrow \{\}$: dictionary for segments in above scan
\STATE $Current\_segs \leftarrow \{\}$: dictionary for segments in current scan
\STATE $current\_segments \leftarrow []$
\STATE $next\_segments \leftarrow []$
\STATE $global\_label \leftarrow 0$
\STATE $Current\_segs \leftarrow find\_segment(scans(1), H_d)$
\STATE $Global\_segs = Current\_segs$
\STATE $Above\_segs = Current\_segs$
\FOR{ $i$ = $2:len(scans)$}
    \STATE $Current\_segs \leftarrow find\_segments(scans(i), H_d)$
    \STATE $len\_current \leftarrow length(Current\_segs)$
    \STATE $temp\_dict = \{\}$
    \FOR{ $j$ = $1:len\_current$}
        \STATE $connect\_index = find\_connection(Current\_segs(j),$ $
        Above\_segs, V_d, mini\_points)$
        \STATE $len\_connect = length(connect\_index)$
        \IF{$len\_connect == 0$}
            \STATE $global\_cluster = global\_cluster + 1$
            \STATE $Global\_segs[global\_label] = Current\_segs(j)$
            \STATE $temp\_dict.[global\_label] = Current\_segs(j)$
        \ENDIF
        \IF{$len\_connect == 1$}
            \STATE $Global\_segs[connect\_index(0)].append(Current\_segs(j))$
            \STATE $temp\_dict[connect\_index(0)] = Current\_segs(j)$
        \ENDIF
        \IF{$len\_connect >= 2$}
            \STATE $mini\_key = fetch\_mini\_key(connect\_index)$
            \STATE $Global\_segs = merge\_delete\_global(Global\_segs,$ $mini\_key, connect\_index)$
            \STATE $Above\_segs = merge\_delete\_above(Above\_segs,$ $mini\_key, connect\_index)$
            \STATE $temp\_dict = merge\_delete\_current(temp\_dict,$ $mini\_key, connect\_index)$
        \ENDIF
    \ENDFOR
    \STATE $Above\_segs = temp\_dict$
\ENDFOR
\end{algorithmic}
\end{adjustwidth}
\end{algorithm}

\section{Occlusion labelling algorithm}\label{appendix:occlusion_label}
\begin{algorithm}[H]
\begin{adjustwidth}{-0.2cm}{}
\caption{Output occlusion label $Occlusion$}
\label{alg:occlusion_label}
\begin{algorithmic}
\STATE Legend:
\STATE
\begin{tabular}{p{0.9cm}p{7cm}}
$\theta\_t$: &angle difference threshold to decide whether two adjacent clusters are occluded or not; \\
$clusters$: &all the clusters which are used to be labeled with occlusion or not;\\
$get\_all\_angle()$:\\ &function to calculate all the angle with respective to original point; \\
$calculate\_range()$:\\ &function to calculate the mean range of one cluster; \\
$calculate\_angle\_overlap$:\\ &function to build the $n*n$ symmetric matrix about angle overlap; \\
$calculate\_diff\_range$:\\ &function to build the $n*n$ symmetric matrix about range difference; \\
\hline
\end{tabular}
\STATE $all\_angles \leftarrow []$: the list of left and right most angle of every cluster
\STATE $all\_ranges \leftarrow []$: the list of range of every cluster
\FOR{ c in $clusters$}
    \STATE $leftmost\_ang, rightmost\_ang = get\_all\_angle(c)$
    \STATE $all\_angles.append([leftmost\_ang-\theta\_t, rightmost\_ang+\theta\_t])$
    \STATE $r = calcualte\_range(c)$
    \STATE $all\_ranges.append(r)$
\ENDFOR
\STATE $occlusion \leftarrow []$
\FOR{ $i$ = $1:len(clusters)$}
    \STATE $ind = find\_angle\_overlap(i,all\_angle)$ %
    \STATE $ind.remove(i)$
    \IF{$len(ind)$==0}
        \STATE $occlusion.append(0)$
    \ELSE
        \FOR{ $j$ in $ind$}
            \IF{$all\_ranges[i] <  all\_ranges[j]$}
                \STATE $occlusion.append(0)$
            \ELSE
                \STATE $occlusion.append(1)$
            \ENDIF
        \ENDFOR
    \ENDIF
\ENDFOR
\end{algorithmic}
\end{adjustwidth}
\end{algorithm}

\section{Particle swarm optimization}\label{append:pso}
\begin{algorithm}[H]
\begin{adjustwidth}{-0.2cm}{}
\caption{Output global best $g^{*}$ and its particle $X^{*}$}
\label{alg:pso}
\begin{algorithmic}
\STATE Legend:
\STATE
\begin{tabular}{p{2.3cm}p{5.5cm}}
$Num\_generation$: &number of iterations; \\
$Num\_particles$: &number of particles; \\
$Range\_particles$: &domain of particle search \\
$Recall\_func()$: &recall calculation function; \\
$Init\_func()$: &initialization function for all particles; \\
$Prop\_generation\_func()$: \\ &proposal generation function; \\
\hline
\end{tabular}
\STATE // {\bf Initialization}
\STATE $X=[x_{1}, x_{2}..., x_{n}]^{T}= Init\_func(Range\_particles)$: randomly initialize all $n$ particles with the defined domain
\STATE $V=[v_{1}, v_{2}..., v_{n}]^{T} = Init\_func(1)$: randomly initialize velocities for $n$ particles
\STATE $g^{*}$ = $Real\_large\_number$: initialize the global best with a large number
\STATE $X^{*} = [x^{*}_{1}, x^{*}_{2}..., x^{*}_{n}]$ = $Real\_large\_number$: initialize the global best with a large number $Real\_large\_value$ 
    \FORALL{$t=0:Num\_generation$}
        \FOR{$i=0:Num\_particles$}
            \STATE $proposals= Prop\_generation\_func(x^{t}_{i})$
            \STATE $recall = Recall\_func(proposals)$
            \STATE // {\bf Find the personnel and global best particle}
            \IF{$recall<x^{*}_{i}$} 
                \STATE $x^{*}_{i} = recall$
            \ENDIF
            \IF{$recall<g^{*}$} 
                \STATE $g^{*} = recall$
            \ENDIF
        \ENDFOR
        
        \STATE // {\bf Particle updating step}
        \FOR{$i=0:Num\_particles$}
                \STATE $v^{t+1}_{i} = \alpha\cdot v^{t}_{i} + \lambda\cdot r_{1}\cdot (g^{*} - x^{t}_{i}) + \theta\cdot r_{2}\cdot (x^{*}_{i} - x^{t}_{i})$
                \STATE $x^{t+1}_{i} = x^{t}_{i} + v^{t+1}_{i}$   
        \ENDFOR
        \STATE // {\bf Particle repair}
        
        \FOR{$i=0:Num\_particles$}
            \IF{$x^{t+1}_{i}$ is out of range}
                \STATE $x^{t+1}_{i}=Init\_func(Range\_particles)$
            \ENDIF
        \ENDFOR
    \ENDFOR 
\end{algorithmic}
\end{adjustwidth}
\end{algorithm}

\end{document}